\documentclass[letterpaper]{article} 
\usepackage{aaai23}  
\usepackage{times}  
\usepackage{helvet}  
\usepackage{courier}  
\usepackage[hyphens]{url}  
\usepackage{graphicx} 
\usepackage{natbib}  
\usepackage{caption} 
\usepackage{algorithm}
\usepackage{algorithmic}
\usepackage{xcolor}

\usepackage{newfloat}
\usepackage{listings}
\DeclareCaptionStyle{ruled}{labelfont=normalfont,labelsep=colon,strut=off} 
\floatstyle{ruled}
\newfloat{listing}{tb}{lst}{}
\floatname{listing}{Listing}
\title{SeedBERT: Recovering Annotator Rating Distributions from an Aggregated Label}
\author {
    Aneesha Sampath,\textsuperscript{\rm 1}
    Victoria Lin, \textsuperscript{\rm 1}
    Louis-Philippe Morency \textsuperscript{\rm 1}
}
\affiliations {
    \textsuperscript{\rm 1} Carnegie Mellon University \\
    \{aneeshas, vlin2, morency\}@cs.cmu.edu
}

\usepackage{bibentry}

\begin{document}

\maketitle

\begin{abstract}
Many machine learning tasks---particularly those in affective computing---are inherently subjective. When asked to classify facial expressions or to rate an individual's attractiveness, humans may disagree with one another, and no single answer may be objectively correct. However, machine learning datasets commonly have just one ``ground truth" label for each sample, so models trained on these labels may not perform well on tasks that are subjective in nature. Though allowing models to learn from the individual annotators' ratings may help, most datasets do not provide annotator-specific labels for each sample. To address this issue, we propose SeedBERT, a method for recovering annotator rating distributions from a single label by inducing pre-trained models to attend to different portions of the input. Our human evaluations indicate that SeedBERT's attention mechanism is consistent with human sources of annotator disagreement. Moreover, in our empirical evaluations using large language models, SeedBERT demonstrates substantial gains in performance on downstream subjective tasks compared both to standard deep learning models and to other models that account explicitly for annotator disagreement.
\end{abstract}




\section{Introduction}
\textit{Subjectivity uncertainty} is uncertainty that arises from data with subjective or ambiguous labels---labels on which human annotators themselves can disagree. Many affective computing tasks, including automatic emotion recognition and hate-speech detection, are highly subjective due to their heavy reliance on human opinion. Different people may perceive a speaker's toxicity level or emotional intent differently. If asked to rate a subjective sample, a group of five annotators may very well provide five different labels.

In contrast, machine learning datasets usually provide a single label for each sample. Labels for machine learning tasks are often crowd-sourced through survey platforms, where researchers request multiple annotators to rate each sample. Prior to a dataset's public release, however, these individual annotators' ratings are usually aggregated into one ``gold standard" label. Flattening annotations into a single label results in loss of information about the subjectivity of the task, and consequently, models trained on aggregated labels may not be able to achieve their desired level of performance on subjective tasks. This raises concerns for 
tasks such as emotion recognition, where perception is highly dependent on personal opinion \cite{variab, deconv, hardtosoft}.

To overcome such concerns and to better represent the nature of subjective tasks, machine learning models must be able to represent the subjectivity of a sample. In this paper, we propose models that can infer subjectivity as a distribution of possible human opinions. Due to the single-label nature of most standard machine learning datasets, we focus our efforts explicitly on models that are able to reconstruct these distributions \textit{from single-label data}. In our modeling mechanism, we hypothesize that annotator subjectivity arises in significant part as a function of the portion of the sample on which the annotator decides to focus.


\paragraph{Illustrated Example.} 
In the remainder of this paper, we take emotion recognition to be an example of a task with subjectivity uncertainty. Many existing emotion recognition datasets, including the language datasets CARER \cite{carer} and Stanford Sentiment Treebank (SST) \cite{sst} provide only one emotion label.

 

We consider two sentences from the SST dataset with similar sentiment scores. In our first example, ``You walk out of The Good Girl with mixed emotions -- disapproval of Justine combined with a tinge of understanding for her actions," the sentiment rating is $0.61$, or slightly positive. The speaker expresses both negative and positive attitudes towards the film---``disapproval of Justine" and ``a tinge of understanding for her actions," respectively---and even explicitly notes mixed emotions in the sentence. These contrasting sentiments, however, are aggregated into just one score that leans positive (where the individual annotators might have considered the sentence to be very negative, slightly negative, very positive, etc.). On the other hand, in another sample from the dataset, ``A smart, steamy mix of road movie, coming-of-age story and political satire," the speaker expresses only positive emotions (indicating that the individual annotators all considered the sentence to be positive), but the sentiment score is $0.67$, which similar to that of the first example. 

These score similarities demonstrate how the use of aggregated labels results in the loss of a nuanced representation of the sentiment. Models trained only on aggregated scores may not learn a sentiment representation as well as those that are trained on the full distribution of annotator ratings.


\paragraph{Contributions.} 

This paper's contributions are threefold:
\begin{itemize}
    \item We propose a deep language model (\textbf{SeedBERT}) that is capable of recovering original annotator distributions from a \textit{single ground-truth label}. This model uses the recovered distribution to predict a label for the task.

    \item We introduce a new set of annotations (\textbf{MOSI-Subjectivity}) of human subjectivity and the mechanisms behind annotator disagreement. These annotations extend the CMU-MOSI sentiment analysis dataset \cite{mosi}.
    \item We present a detailed evaluation with both empirical performance analysis and a study of human agreement.
\end{itemize}

\section{Related Work}

\paragraph{Inter-Annotator Disagreement.} 
Annotator disagreement can  arise when tasks are inherently subjective. Affective semantics are particularly difficult to model since people's biases, experiences, and knowledge lead them to form different opinions about the same data. This leads to performance degradation on tasks involving expressive language \cite{subjannotations}. In subjective contexts, it can therefore be useful to move away from a single ground-truth label and to instead acknowledge that multiple answers may be valid. 

Several prior works attempt to address this unmet need by explicitly modeling individual annotators. Models such as Jury Learning \cite{jurylearning} and HuBi-Medium \cite{hubi} propose joint training of text embeddings and annotator embeddings to allow for predictions on the individual annotator-level. Although these approaches move toward personalized annotator modeling, they require not only that multiple annotators' labels are available for each data sample, but also that annotator-identification information is available. This requirement contrasts with the typical crowd-sourced machine learning datasets, in which annotator-level information is not provided.


\paragraph{Modeling Perception Uncertainty.} An additional body of work attempts to account for subjectivity by modeling perception uncertainty directly. \citet{hardtosoft} propose a multi-task framework that simultaneously predicts an emotion label and an estimate of annotator disagreement. The resulting model outperforms single-task emotion recognition models.
\citet{dealing} also propose a multi-task framework---one in which each task consists of modeling an individual annotator. The variance of the resulting annotator distribution is taken as the inter-annotator disagreement measure. However, both of these methods require access to annotator-level labels, which are not commonly available in public datasets, at training time. Moreover, the method proposed by \citet{dealing} requires that each annotator label a large portion of the data.

\section{MOSI-Subjectivity Dataset}

To gain a better understanding of subjectivity uncertainty and the mechanisms by which it arises, we collected extensive human ratings of sentiment and subjectivity on an emotion recognition task. We used these ratings to evaluate the empirical performance and human agreement of our proposed method, SeedBERT.

\paragraph{Dataset.}
We used video clips\footnote{We initially collected data using transcripts only, but obtained poor inter-rater reliability on all questions.} from the Multimodal Corpus of Sentiment Intensity dataset (CMU-MOSI), a collection of over 2000 YouTube movie reviews with opinion-level annotations for sentiment and subjectivity. 

\paragraph{Data Collection.}

We collected new labels for 500 randomly selected audiovisual CMU-MOSI samples using the \textit{Prolific.co}\footnote{\url{https://www.prolific.co/}} crowdsourcing platform. For each sample, we asked five high-quality annotators (approval rating $\geq 98\%$) to answer the following questions.

\begin{enumerate}
    \item Is the speaker expressing \textit{any} emotion in the video clip? 
    \item Select the emotion \textit{most present} in the video. 
    \item Is the person expressing multiple emotions in the video? If so, select \textit{all of the emotions present} in the video.
\end{enumerate}

We intended the first question---which has relatively little ambiguity---to be a calibration question in order to establish the expected level of inter-annotator agreement in the absence of any subjectivity. This question was answered using a binary rating scale, with options \textit{Yes} and \textit{No}.

The second question, which requests that annotators mark the most salient emotion in the video clip, is one commonly asked in emotion recognition data collection procedures. CMU-MOSEI \cite{mosei}, for instance, used a similar question to collect its labels. As with CMU-MOSEI, the six Ekman emotions were presented as label options in a multiple-choice format.

The third question is a multiple-select question in which annotators were instructed to select all of the Ekman emotions that the speaker expresses in the video clip. This question allows raters to express greater nuance on the emotion recognition task rather than limiting them to a single response, which we felt could have implications for the subjectivity of the task.

After collecting a portion of the annotations and examining them, we hypothesized that disagreement might arise due to annotators attending to different parts of the sentence, where each part may emphasize a different emotion. To verify this hypothesis, we collected additional annotations for another $60$ samples, wherein we asked raters to explain their selections. Again recruiting five high-quality annotators (approval rating $\geq 98\%$) to rate each sample, we asked:

\begin{enumerate}
    \setcounter{enumi}{3}
    \item On the previous page, you selected $X$ as the emotion most present in the clip. Which words or phrases led you to your answer? Please highlight the portions of the transcript that led you to make your selection.

\end{enumerate}

\subsection{Analysis of Annotator (Dis)agreement}
\paragraph{Quantitative Analysis.}
To determine the level of agreement among annotators for each of the questions asked, we calculated the Cohen's kappa coefficient \cite{kappa} using the \textit{agreement} software package.\footnote{\url{https://github.com/jmgirard/agreement}} To account for the multiple-selection setting of the third question, we treated the responses to each emotion as answers to a single binary question, then averaged Cohen's kappa among all emotions.

\begin{table}[h!]
    \centering
    \begin{tabular}{p{6.6cm} p{0.8cm}}
    \hline
    \textbf{Question} & \textbf{Kappa} \\
    \hline 
    Is the speaker expressing \textit{any} emotion in the video? & $0.744$ \\
    Select the emotion \textit{most present} in the video. & $0.410$ \\
    Select \textit{all of the emotions present} in the video. & $0.710$ \\
    \end{tabular}
    \caption{Cohen's kappa coefficients for unordered categories for the three survey questions.}
    \label{table:icc}
\end{table}

We observe that annotators exhibit similar levels of agreement for the calibration question and the multi-label question (Table \ref{table:icc}). However, agreement is substantially lower for the second question concerning the emotion most present in the sample. These results seem to indicate that annotators can acknowledge and agree upon the presence of other opinions, but greater subjectivity and disagreement arise when annotators are asked to select the most salient opinion.

\paragraph{Qualitative Analysis.} 
Observing this discrepancy, we sought to investigate why annotators have high agreement when selecting all emotions present but low agreement when selecting the primary emotion. In Table \ref{table:qual}, we provide selected examples to illustrate our analysis.

\begin{table}[h!]
    \centering
    \begin{tabular}{|| p{5cm} p{2.4cm} ||}
    \hline
    \textbf{Sample} & \textbf{Main Emotions} \\
    \hline \hline
    these action sequences are certainly \textit{a lot better} than the \textbf{incoherent impossible to follow slop} that we see in a typical michael bay movie & \textit{Happiness}, \textbf{Anger}, \textbf{Disgust} \\
    \hline
    and i honestly think \textbf{if we hadnt have seen watchman} this would be the \textit{best credit sequence} of the year & \textit{Happiness}, \textbf{Sadness}, \textit{Surprise} \\
    \hline
    but there are you know \textbf{there're critics of movies that are sometimes too thoughtful} in a way and \textit{i think that would never let me go} & \textbf{Disgust}, \textit{Surprise} \\
    \hline
    \end{tabular}
    \caption{Selected examples from the first survey. The first column is the transcript of the video clip shown to the annotators, and the second column is a set of the emotions selected as most present by the five annotators.}
    \label{table:qual}
\end{table}

Given the transcript of the sample and the set of annotators' responses to the emotion most present in the sample, we observe that different words or phrases within the sample may elicit different emotions. Taking the first example from Table \ref{table:qual}, we can see that the italicized part of the sentence is characterized by a \textit{happy tone}, whereas the bolded part has a \textbf{disgusted or angry tone}. While all annotators indicated that \textit{happiness} and \textit{disgust} or \textit{anger} were present when answering the third question, they disagreed as to which of the three was the primary emotion when answering the second question. This leads us to hypothesize that \textbf{annotators attend to different parts of the sentence}, which leads to low agreement when selecting the most salient emotion but high agreement when selecting all of the emotions present. 

\begin{figure}[h!]
  \centering
  \includegraphics[scale=0.16]{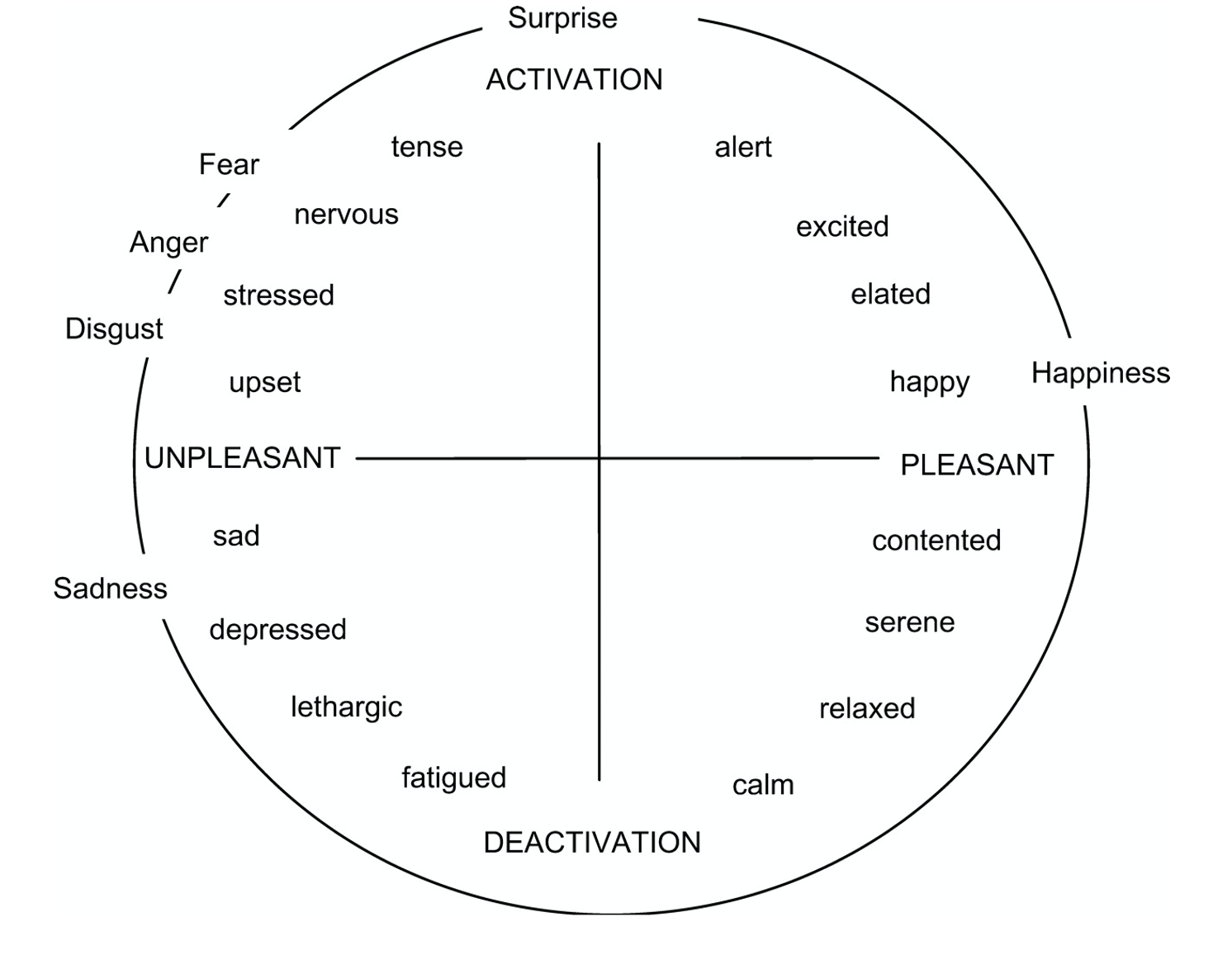}
  \caption{Circumplex model of affect.}
  \label{fig:wheel}
\end{figure}

As mentioned in the Data Collection section, we collected further annotations to verify this hypothesis, this time asking annotators to explain their selections for the most salient emotion by highlighting specific words and phrases in the transcript that led them to their decision. We found that our hypothesis holds when the emotions present are adjacent on the circumplex model of affect (Figure \ref{fig:wheel}), in which similar emotions are adjacent to one another and opposite emotions are across from one another \cite{circumplex}. 

\begin{table}[h!]
    \centering
    \begin{tabular}{|| p{1.5cm} p{3.7cm} p{1.6cm} ||}
    \hline
    \textbf{Emotion} & \textbf{Supporting Evidence} & \textbf{Annotators} \\
    \hline \hline
    Happiness &  a lot better & 2\\
    \hline
    Anger & incoherent impossible to follow slop & 2\\
    \hline
    Disgust & incoherent impossible to follow slop & 1 \\
    \hline
    \end{tabular}
    \caption{Selected example from the second set of surveys. Two annotators selected \textit{Happiness} as the main emotion present, two selected \textit{Anger}, and one selected \textit{Disgust}.}
    \label{table:explain}
\end{table}

Still using the first sample from Table \ref{table:qual} as an illustrative case, we observe in Table \ref{table:explain} that for the non-adjacent emotion pairs (\textit{Happiness}, \textit{Anger}) and (\textit{Happiness}, \textit{Disgust}), annotators cited different portions of the transcript to support their answer for the main emotion present, suggesting that they attend to different parts of the sentence. However, this finding did not hold for adjacent emotions, as the same portion of the sentence is provided as justification for both the \textit{Anger} and \textit{Disgust} labels. \textbf{This result may suggest that subjectivity may arise both from the part of the input attended to and from personal interpretation of the same portion of the input}. For the remainder of this work, we will focus on subjectivity due to the part of the utterance attended to.

\section{Methods}

\begin{figure}[h!]
  \centering
  \includegraphics[scale=0.28]{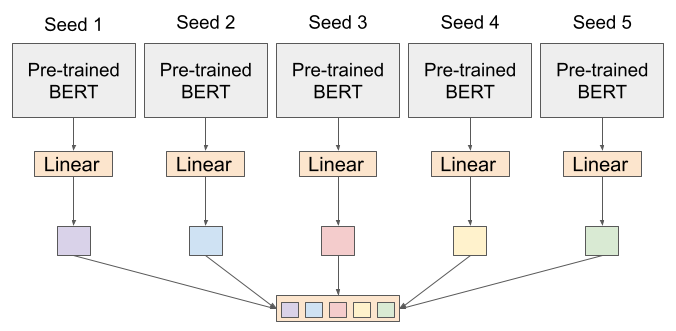}
  \caption{SeedBERT architecture. Pre-trained BERT encoders are paired with a linear layer initialized from a random seed.}
  \label{fig:seedbert}
\end{figure}

\paragraph{SeedBERT.} Our proposed method, SeedBERT (Figure \ref{fig:seedbert}), is premised on the idea that a collection of randomly initialized layer in a neural network will (by virtue of their different starting points) learn to attend to different parts of an input, even when trained on the same data. When the input is inherently subjective, the linear layer---in attending to a particular portion---will represent the opinions and biases of an arbitrary annotator. Following this reasoning, SeedBERT is a collection of $n$ identical networks consisting of pre-trained BERT encoders with a task-specific linear layer. Each network is trained on the same dataset with identical loss functions, with the only difference being that each network has a different random initialization of the final linear layer. Each network is meant to reflect one annotator, and SeedBERT's final prediction is the majority vote of the networks.

As an ensemble of networks, SeedBERT bears certain methodological similarities to classical ensemble methods like bagging \cite{bagging}. Rather than using a collection of weak learners to improve predictive performance, however, SeedBERT leverages a powerful existing model---pre-trained BERT---and uses its ``ensembling" to induce individual learners to attend to different portions of the input.

\paragraph{Evaluation.} We compare our approach, SeedBERT, against four baseline models, two of which we describe previously in the Related Work section. These latter two models are trained on individual annotator ratings, which are not typically available in machine learning datasets. Due to their use of this additional training data, we can consider them to be oracle models that provide upper bounds on the ability of a model to learn annotator distributions. 
All models use pre-trained BERT as an encoder \cite{bert}, with a final linear layer fine-tuned for the task via the Adam optimizer. We use a learning rate of $5 \times 10^{-6}$ and fine-tune for $3$ epochs. We evaluated our models on metrics accuracy and $F_1$-score using 5-fold cross-validation and repeated this process 5 times with different splits of the data.

\paragraph{Baseline Models.} Our first baseline model \textbf{(BERT)} reflects a system where there is no explicit attempt to account for subjectivity. This model consists of a single pre-trained BERT encoder with a task-specific final linear layer.

Our second baseline model incorporates Bayesian neural networks \textbf{(BNN)}, which have previously been used to estimate prediction uncertainty due to their use of probablistic rather than deterministic weights. Recent work has applied BNNs to speech emotion recognition in order to directly estimate annotator distributions \cite{bayes}. We leverage a similar architecture of an encoder followed by a BNN, but use a transformer-based BERT encoder.

We also compare SeedBERT against two oracle models trained on individual annotator ratings, which are not typically available in machine learning datasets. We use a Label Distribution Learning (\textbf{LDL}) framework \cite{ldl} and the \textbf{Multi-Task} framework proposed by \citet{hardtosoft} to determine the upper bounds of achievable model performance when modeling annotator distributions. 

\section{Results \& Discussion}

\begin{figure}[h!]
  \centering
  \includegraphics[scale=0.3]{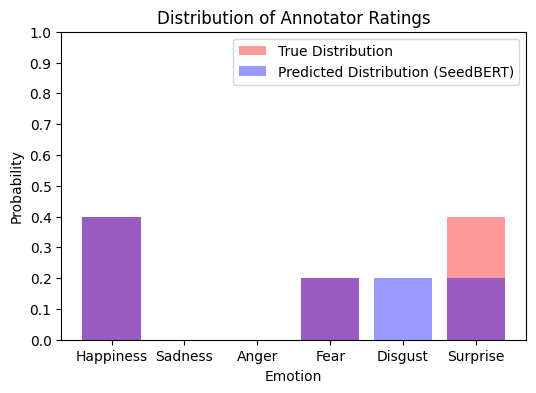}
  \includegraphics[scale=0.3]{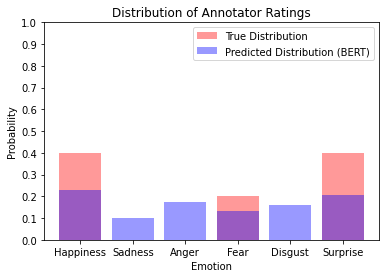}  
  \caption{Example of SeedBERT predicted annotator distribution of emotions (left) and BERT predicted distribution (right). The predicted distributions are in blue, while the true annotator distributions are in red.}
  \label{fig:distributions}
\end{figure}

In our empirical evaluation, we observe in Table \ref{table:results} that SeedBERT model outperforms all other models on the emotion recognition task. Notably, given only the aggregated label at training time, SeedBERT outperforms not only the baselines (standard pre-trained BERT and BNN) but also the two models trained on individual annotator data. These results suggest that SeedBERT's mechanism of reconstructing annotator distributions is effective in improving performance on subjective tasks.

Through qualitative analyses, we find that an individual BERT model tends to predict uniform distributions of annotator ratings (i.e., uniform across emotions) irrespective of the input sample (Figure \ref{fig:distributions}), possibly because it cannot fit well to low-agreement data. As seen in Table \ref{table:results}, this shortcoming has clear negative implications for its performance on the emotion recognition task. On the other hand, the SeedBERT approach moves toward reflecting annotators' biases, as each random initialization can reflect the preferences of an arbitrary annotator. As shown in Figure \ref{fig:distributions}, the annotator distribution predicted by SeedBERT much more closely reflects the true annotator distribution.

We further find that certain random initializations of the individual SeedBERT networks tended to bias toward a subset of the Ekman emotions---much in the same way that a human annotator might have a tendency to favor certain emotions in their ratings. These observations align with our hypothesis that the random initializations may serve as proxies for individual annotator perceptions or biases. 

\begin{table}[h!]
    \centering
    \begin{tabular}{ccc}
    \hline
    \textbf{Model} 
    & \textbf{Accuracy}
    & \textbf{$F_1$-Score}\\
    \hline
    BERT 
    & .204
    & .102\\
    LDL 
    & .284
    & .0821 \\
    Multi-Task
    & .180
    & .0954 \\
    BNN 
    & .150
    & .132 \\
    SeedBERT 
    & \textbf{.308} 
    & \textbf{.183} \\

    \end{tabular}
    \caption{SeedBERT outperforms all baselines when averaged across 5 iterations of 5-fold cross validation.}
    \label{table:results}
\end{table}

\section{Conclusions}

In this paper, we analyze human subjectivity and the mechanisms by which annotator disagreement arises. We propose SeedBERT, a novel method capable of recovering original annotator distributions from a single ground-truth label to ultimately improve performance on downstream tasks. Via SeedBERT, we find that manipulating the random initializations of pre-trained models can serve as an effective proxy for modeling varying annotator perceptions.

Our findings should be considered in light of several points. First, our extension to CMU-MOSI consisted of a relatively small number of samples (500) for fine-tuning a large language model. Second, fewer samples were rated for \textit{anger} and \textit{fear}, which could limit the efficacy of our method when evaluating these emotions. Third, although CMU-MOSI was designed to be a multimodal dataset, we implemented SeedBERT and its comparison methods as unimodal language models, which likely limits their predictive ability.
Fourth, the validation set accuracies and $F_1$-scores are relatively low across all models, which warrants further exploration.
Finally, though random initializations can effectively represent a collection of arbitrary annotators, there are cases where prior information about the annotators' preferences is known (for example, an annotator's favorite movie genre). In these cases, SeedBERT may be improved by incorporating this prior information to represent a \textit{specific} annotator distribution rather than an arbitrary one.


\bibliography{aaai23}

\begin{thebibliography}{17}
\providecommand{\natexlab}[1]{#1}

\bibitem[{Alm(2011)}]{subjannotations}
Alm, C.~O. 2011.
\newblock Subjective natural language problems: Motivations, applications,
  characterizations, and implications.
\newblock In \emph{Proceedings of the 49th Annual Meeting of the Association
  for Computational Linguistics: Human Language Technologies}, 107--112.

\bibitem[{Bagher~Zadeh et~al.(2018)Bagher~Zadeh, Liang, Poria, Cambria, and
  Morency}]{mosei}
Bagher~Zadeh, A.; Liang, P.~P.; Poria, S.; Cambria, E.; and Morency, L.-P.
  2018.
\newblock Multimodal Language Analysis in the Wild: {CMU}-{MOSEI} Dataset and
  Interpretable Dynamic Fusion Graph.
\newblock In \emph{Proceedings of the 56th Annual Meeting of the Association
  for Computational Linguistics (Volume 1: Long Papers)}, 2236--2246.
  Melbourne, Australia: Association for Computational Linguistics.

\bibitem[{Breiman(1996)}]{bagging}
Breiman, L. 1996.
\newblock Bagging predictors.
\newblock \emph{Machine learning}, 24(2): 123--140.

\bibitem[{Cohen(1960)}]{kappa}
Cohen, J. 1960.
\newblock A coefficient of agreement for nominal scales.
\newblock \emph{Educational and psychological measurement}, 20(1): 37--46.

\bibitem[{Davani, D{\'\i}az, and Prabhakaran(2022)}]{dealing}
Davani, A.~M.; D{\'\i}az, M.; and Prabhakaran, V. 2022.
\newblock Dealing with disagreements: Looking beyond the majority vote in
  subjective annotations.
\newblock \emph{Transactions of the Association for Computational Linguistics},
  10: 92--110.

\bibitem[{Geng(2016)}]{ldl}
Geng, X. 2016.
\newblock Label distribution learning.
\newblock \emph{IEEE Transactions on Knowledge and Data Engineering}, 28(7):
  1734--1748.

\bibitem[{Gordon et~al.(2022)Gordon, Lam, Park, Patel, Hancock, Hashimoto, and
  Bernstein}]{jurylearning}
Gordon, M.~L.; Lam, M.~S.; Park, J.~S.; Patel, K.; Hancock, J.; Hashimoto, T.;
  and Bernstein, M.~S. 2022.
\newblock Jury learning: Integrating dissenting voices into machine learning
  models.
\newblock In \emph{CHI Conference on Human Factors in Computing Systems},
  1--19.

\bibitem[{Gordon et~al.(2021)Gordon, Zhou, Patel, Hashimoto, and
  Bernstein}]{deconv}
Gordon, M.~L.; Zhou, K.; Patel, K.; Hashimoto, T.; and Bernstein, M.~S. 2021.
\newblock The Disagreement Deconvolution: Bringing Machine Learning Performance
  Metrics In Line With Reality.
\newblock In \emph{Proceedings of the 2021 CHI Conference on Human Factors in
  Computing Systems}, CHI '21. New York, NY, USA: Association for Computing
  Machinery.

\bibitem[{Han et~al.(2017)Han, Zhang, Schmitt, Pantic, and
  Schuller}]{hardtosoft}
Han, J.; Zhang, Z.; Schmitt, M.; Pantic, M.; and Schuller, B. 2017.
\newblock From Hard to Soft: Towards More Human-like Emotion Recognition by
  Modelling the Perception Uncertainty.
\newblock In \emph{Proceedings of the 25th ACM International Conference on
  Multimedia}, MM '17, 890–897. New York, NY, USA: Association for Computing
  Machinery.

\bibitem[{Kenton and Toutanova(2019)}]{bert}
Kenton, J. D. M.-W.~C.; and Toutanova, L.~K. 2019.
\newblock BERT: Pre-training of Deep Bidirectional Transformers for Language
  Understanding.
\newblock In \emph{Proceedings of NAACL-HLT}, 4171--4186.

\bibitem[{Koco{\'n} et~al.(2021)Koco{\'n}, Gruza, Bielaniewicz, Grimling,
  Kanclerz, Mi{\l}kowski, and Kazienko}]{hubi}
Koco{\'n}, J.; Gruza, M.; Bielaniewicz, J.; Grimling, D.; Kanclerz, K.;
  Mi{\l}kowski, P.; and Kazienko, P. 2021.
\newblock Learning personal human biases and representations for subjective
  tasks in natural language processing.
\newblock In \emph{2021 IEEE International Conference on Data Mining (ICDM)},
  1168--1173. IEEE.

\bibitem[{Raj~Prabhu et~al.(2022)Raj~Prabhu, Carbajal, Lehmann-Willenbrock, and
  Gerkmann}]{bayes}
Raj~Prabhu, N.; Carbajal, G.; Lehmann-Willenbrock, N.; and Gerkmann, T. 2022.
\newblock End-To-End Label Uncertainty Modeling for Speech-based Arousal
  Recognition Using Bayesian Neural Networks.
\newblock 151--155.

\bibitem[{Russell(1980)}]{circumplex}
Russell, J.~A. 1980.
\newblock A circumplex model of affect.
\newblock \emph{Journal of personality and social psychology}, 39(6): 1161.

\bibitem[{Saravia et~al.(2018)Saravia, Liu, Huang, Wu, and Chen}]{carer}
Saravia, E.; Liu, H.-C.~T.; Huang, Y.-H.; Wu, J.; and Chen, Y.-S. 2018.
\newblock {CARER}: Contextualized Affect Representations for Emotion
  Recognition.
\newblock In \emph{Proceedings of the 2018 Conference on Empirical Methods in
  Natural Language Processing}, 3687--3697. Brussels, Belgium: Association for
  Computational Linguistics.

\bibitem[{Socher et~al.(2013)Socher, Perelygin, Wu, Chuang, Manning, Ng, and
  Potts}]{sst}
Socher, R.; Perelygin, A.; Wu, J.; Chuang, J.; Manning, C.~D.; Ng, A.; and
  Potts, C. 2013.
\newblock Recursive Deep Models for Semantic Compositionality Over a Sentiment
  Treebank.
\newblock In \emph{Proceedings of the 2013 Conference on Empirical Methods in
  Natural Language Processing}, 1631--1642. Seattle, Washington, USA:
  Association for Computational Linguistics.

\bibitem[{Zadeh et~al.(2016)Zadeh, Zellers, Pincus, and Morency}]{mosi}
Zadeh, A.; Zellers, R.; Pincus, E.; and Morency, L.-P. 2016.
\newblock Multimodal sentiment intensity analysis in videos: Facial gestures
  and verbal messages.
\newblock \emph{IEEE Intelligent Systems}, 31(6).

\bibitem[{Zhang, Essl, and Mower~Provost(2017)}]{variab}
Zhang, B.; Essl, G.; and Mower~Provost, E. 2017.
\newblock Predicting the Distribution of Emotion Perception: Capturing
  Inter-Rater Variability.
\newblock In \emph{Proceedings of the 19th ACM International Conference on
  Multimodal Interaction}, ICMI '17, 51–59. New York, NY, USA: Association
  for Computing Machinery.

\end{thebibliography}

\end{document}